\documentclass{article}
\usepackage[utf8]{inputenc}
\usepackage{amsmath} 
\usepackage[utf8]{inputenc}
\usepackage{graphicx}
\usepackage{csquotes}
\usepackage[english]{babel}
\usepackage{hyperref}
\usepackage{amssymb}
\usepackage{amsthm}
\newenvironment{claim}[1]{\par\noindent\underline{Claim:}\space#1}{}

\newenvironment{claimproof}[1]{\par\noindent\underline{Proof:}\space#1}{\hfill$\blacksquare$}
\newenvironment{lemma}[1]{\par\noindent\underline{Lemma:}\space#1}{}
\newenvironment{lemmaproof}[1]{\par\noindent\underline{Proof:}\space#1}{\hfill$\blacksquare$}

\newenvironment{theorem}[1]{\par\noindent\underline{Theorem:}\space#1}{}
\newenvironment{theoremproof}[1]{\par\noindent\underline{Proof:}\space#1}{\hfill$\blacksquare$}

\newenvironment{definition}[1]{\par\noindent\underline{Definition:}\space#1}{}

\usepackage[utf8]{inputenc}
\usepackage{graphicx}

\graphicspath{{plot_fold/}}
 \usepackage[
backend=biber,
style=alphabetic,
sorting=ynt
]{biblatex}
\title{Goodness of fit metrics for Multi-class Predictor  }
\usepackage{listings}
\author{Uri Itai and Natan Katz }

\begin{document}
\date{May  2022}https://www.overleaf.com/project/61b73c15e016b51a702376c0

%  \usepackage[
% backend=biber,
% style=alphabetic,
% sorting=ynt
% ]{biblatex}
% \addbibresource{sample.bib}

\maketitle

\begin{abstract}
The multi-class prediction had gained popularity over recent years.
Thus measuring fit goodness becomes a cardinal question that researchers often have to deal with. Several metrics are commonly used for this task. However, when one has to decide about the right measurement, he must consider that different use-cases impose different constraints that govern this decision. A leading constraint at least in \emph{real world} multi-class problems is imbalanced data: Multi categorical problems hardly provide symmetrical data. Hence, when we observe common KPIs (key performance indicators), e.g., Precision-Sensitivity or Accuracy, one can seldom interpret the obtained numbers into the model's actual needs. 
We suggest generalizing Matthew's correlation coefficient into multi-dimensions. This generalization is based on a geometrical interpretation of the generalized confusion matrix.
\noindent

\end{abstract}
\section{Introduction}
Measuring the performance of a classifier plays a major role in many aspects of machine learning. Methods to grad classifier are called  \textbf{the goodness of fit}.
The goodness of fit for a binary classification had attracted attention in the field of machine learning and statistics \cite{anderson1954test}. Yet, the classification of more than two classes (multi-class) had gotten less research focus. Moreover, in many cases when such a setting occurs many researchers are using ad-hoc hacks. These methods usually fail to give a full description of the multi-class nature of the goodness of the fit. In this paper, we will provide metrics for the goodness of fit for multi-class problems which take into account the essence of the setting. In addition, these indicators are sensitive to the absence of natural order. Thus, we will start with the presentation of what the Multi-Class Problem is. 

\subsection{What is a Multi-Class Problem? }
Most machine learning applications provide a binary framework. For example disease detection, or bi-object detection. 
The outcome of these problems is always isomorphic to a Boolean state: \emph{True} or \emph{False}.
Following this, one can easily define the plausible errors (False positive and False-negative). The goodness of fit is merely a statistic over these errors. Nonetheless, when the number of classes exceeds the two, this isomorphism breaks. This takes place due to the definition of the error. This entails that the above needs to be modified. 
We denote prediction problems with more than two possible outcomes \textbf{multi-class} prediction problems.  \footnote{In this paper, multi-class means that there exist at least three classes.}

Approach to handle this:
\begin{itemize}
    \item  \emph{One-to-many} manners. It is often used, in cyber attacks.
    \item In some problems different types of errors have different costs. 
    \item Assume the classes are embedded in some Euclidean space. Consequently, we have a regression problem in a latent space. Nevertheless, it is not grantee that this is possible. Moreover, even so, finding the right embedding into the latent space is a difficult task. 
\end{itemize}

  The above assumes some kind of topological order of the classes. Namely, hierarchy, latent space, etc. In this work, the analysis does not take any of this into account.

\subsection{The Curse of imbalanced data}
Many classification algorithms make assumptions about the data prevalence. 
However, in many real-life prediction problems, the data classes are imbalanced. Namely, the magnitude of the classes differs. Such traffic distribution may provide good results accuracy-wise but allow severe weaknesses, The following example may clarify this claim:
\subsubsection{Imbalanced Data - Numerical Example}
Consider a Covid-19 test. Its accuracy is 99 percent.  
In a certain clinic, the medical staff uses this test for its 50 thousand visitors. Assume that a thousand of them are sick (indeed an extremely high rate). The clinic's tests will detect 990 of them. Still, it will declare five hundred healthy visitors as \emph{sick} despite they are not. Namely one out of three detected \emph{sick} is healthy.
This phenomenon takes place since the predictor always prefers the dominant class.
\subsection*{}
Such scenarios may occur in other disciplines such as bio-metrics or cyber.
To be able to overcome this we define the confusion matrix.

\subsection{Confusion Matrix} 

As we learned that accuracy can be misleading, we need to search for different indicators. To overcome this the confusion matrix may be considered a natural choice  

\[
\begin{pmatrix}
TP& FP\\
FN&TN
\end{pmatrix} . 
\] 
Where 
 
\begin{itemize}
    \item TP - The number of true positive
    \item FP -  The number of false positive
    \item TN -  The number of true negative 
    \item FN -  The number of false negative.
\end{itemize}

Given this matrix, our objective is to find a scalar function that processes this matrix and outputs a good fit measurement.  Four more useful measures are
\begin{itemize}
    \item Precision- The probability of TP given the predicted positive. 
    \item Sensitivity- the probability of TP given the being positive (in some papers it is called Recall). 
    \item Specificity - the probability of TN given the being negative.
    \item NPV (negative predictive value) - the probability of TN given the predicted negative.  
\end{itemize}
We will discuss the following metrics:

\begin{itemize}
    \item F1-score
    \item AUC
    \item Matthews correlation coefficient.
\end{itemize} 
 In this paper, we assume the general case of multi-class. In plain English, we do not have one preferable class. Furthermore, we cannot assume any order of any type. I.e., no topological structure between the classes is known to the user.
In this paper, we will demonstrate methods to construct classifiers for the multi-classes case.
This paper is arranged the following: in Section \ref{Sec:twoclass} we will introduce metrics for two classes. Concerning the geometrical properties. The requirement for multi-class would be discussed in Section \ref{Sec:multi}. In addition, we will provide examples of such metrics.
The main contribution, of this paper, is in this section. Code for Python would be supplied in Section \ref{sec:code}. We will end with closing remarks and acknowledgments in Section \ref{sec:clos}.

\section{Metrics for two dimensions}\label{Sec:twoclass}
Assume a binary prediction problem. We developed a model, ran an evaluation test, and obtained a confusion matrix. Now we wish that our fit goodness indicator will reflect the strength of our model. Intuitively thinking, we expect our model to be considered \emph{good} for any choice of the label set. However, this is not the case: some models may detect better one class than the other. For imbalanced data, these gaps can become essential. Namely, the observation of whether a model is good depends on the target's labeling: which class is defined as "True". 
\subsection{F1-score} 
F1-score is often used in \emph{real world} binary prediction problems. In many machine learning projects it is the \emph{first choice of metric} for goodness of fit score.  F1-score is the Harmonic mean of the Sensitivity and the Precision \footnote{ An equivalent definition is 
\begin{displaymath}
F1=\frac{TP}{TP+\frac{1}{2}(FP+FN)} 
\end{displaymath}} .
 
\begin{equation}\label{EQ:harmonic_mean}
H(a,b) =  \left\{\begin{aligned}
                \frac{2}{\frac{1}{a} +\frac{1}{b} } =  \frac{1}{A(\frac{1}{a},\frac{1}{b})}&   & \text{for }a, b > 0 \\
                0 &   & \text{otherwise.}
               \end{aligned}\right. 
    \end{equation}

Where $A(a,b)$ is the algebraic mean ($A(a,b)= \frac{a+b}{2} $). 
Namely, it produces a hierarchy between the positive class and the negative class.  Namely, the F1-score  \textbf{is not invariant for the binary case}.
For a bigger number of classes we may suffer from both:
\begin{itemize}
    \item The need to extend the definition of errors.
    \item Combinatorially, the sensitivity to class ordering increases. 
\end{itemize}

The geometric mean for non-negative pair of number is:
\begin{equation}\label{EQ:geometric_mean}
G(a,b) =\sqrt{a*b} . 
\end{equation}

 The inequality means states
 \begin{equation}\label{EQ:mean_inequ}
 \min(a,b) \le H(a,b)\le G(a,b)\le A(a,b) \le \max(a,b) .
 \end{equation} The equality holds if and only if $a=b$. In addition, Equation (\ref{EQ:mean_inequ}) holds for defining the means for $k$ variable, $k\in \mathbb{N}$. For this case the means would be define the

 following:\[ A(\{a_i\}_{i=1}^k) = \frac{\sum_{i=1}^k a_i}{k} \] 
 \[\quad  G(\{a_i\}_{i=1}^k) = \sqrt[k]{\prod_{i=1}^k a_i}  \]
 and \[ H(\{a_i\}_{i=1}^k) = (A(\{a_i^{-1}\}_{i=1}^k)^{-1}\]

 We can obtain a good explanation for growing the popularity of the F1-score for two classes. Replacing the Harmonic mean to Geometric mean one gets the \cite{fowlkes1983method}.
 \\Using (\ref{EQ:mean_inequ}) we have  $ F1 \le FM$, where $ F1$ is the F1-score, $FM$ is the Fowlkes–Mallows index. It is noteworthy that if one flips the positive and the negative classes, a different score might appear. This is a major drawback in high dimensions. In other words, if the predictor favors one class, the metric would not point this out. To overcome this, one can define the above F1-score as the F1-1-score. In addition, defining F1-0-score as the harmonic mean of the specificity and the sensitivity. F1-score and the F1-0-score do not solve the hierarchy problem. However, taking any 
 \emph{reasonable average} of these quantities would. This is would construct a \emph{blind to a bias of the predictor}. To better formulate this we define \emph{reasonable average}:
 \begin{definition}
Let $S:\mathbb{R}^{+,n} \xrightarrow{} \mathbb{R}^{+}$ be a
 multi-dimension non-negative valued differential strictly monotone (in each component). In addition for each permutation $\pi(\cdot)$
 \[S(x_1,x_2,\ldots x_n) = S(\pi(x_1,x_2,\ldots x_n)) . \]If it agrees with
 \[
 \min_{0 \le i \le n}(x_i)\le S(x_1,x_2,\ldots x_n)\le  \max_{0 \le i \le n}(x_i)  ,
 \]then it is consider a \textbf{reasonable average}. 

 \end{definition}
 We note that all the means that we define above are reasonable averages. \\
 It is easy to see that for the flat vector, $(x,x,\ldots x)$ for fixed non-negative value $x$, we have $S(x,x,\ldots x)=x$. \\
  For more details on average functions see \cite{sharon2013approximation}.
  Additional observation is the consider the $\ell^p$ average for positive $a,b$:
  \begin{equation}\label{EQ:lp}
  \ell^p(a,b) = \left\{\begin{aligned}
                \left(\frac{a^p+b^p}{2}\right)^p&   & \text{for } p \ne 0  \\
                G(a,b) &   &  p = 0.\end{aligned} \right.
  \end{equation}

  It is well known that $ \ell^p(a,b)$ is continuous in $a,b$ and $p$. Moreover,  $\ell^p(a,b)$ is monotonic increasing in $p$. Moreover, for $p =1$ we get the algebraic mean and for $p=-1 $ we get the harmonic mean. If one let $p\rightarrow{}\infty$ we would get the maximum and for $p\rightarrow{}-\infty$ we get the minimum. Another interesting case is $a$ is the sensitivity and $b$ is precision for $p=-1$ we get the F1-score and for $p=0$ we get the 
  Fowlkes–Mallows index. Following this one can construct any continuous goodness of fit metric as close as possible to the minimum of the sensitivity and precision. 
  
  Equation \ref{EQ:lp} can be generalize to $n$ as the following:
   \begin{equation}\label{EQ:lpn}
  \ell^p(a,b) = \left\{\begin{aligned}
                \left(\frac{\sum_{i=1}^n a_i^p+}{n}\right)^p&   & \text{for } p \ne 0  \\
                G(\{a_i\}_{i=1}^n) &   &  p = 0. \end{aligned}\right.
  \end{equation}

  Plugging the sensitivity, precision, specified and the NPV in \ref{EQ:lpn} for $p\le1$ we get additional goodness for fit metric.  
 \subsection{AUC}
AUC -\textbf{Area Under the Curve} is a common \emph{goodness of fit} estimator as well. It measures the area under ROC curve(FP vs TP normalized by the number of trials). Although it is commonly used, it does not have a natural definition for multi-classes problems and it is extremely sensitive to imbalanced data as the numbers of FP and TP are not scalable. 
Similarly to F1-score, there is a hierarchy between the positive and negative classes. 

\subsection{Matthew's Correlation Coefficient}
\subsubsection{Preliminary}
Chico and Jurman in \cite{chicco2020advantages}
had discussed the advantages of the Matthews correlation coefficient over F1-score and accuracy in the case of binary classification evaluation. The main advantage of this measure is that it detects whether the prediction is biased against (or for) some classes. In addition, we will show that the Matthews correlation coefficient is invariant for renaming classes or flipping the target and the predicted vector. This helps to tackle the imbalanced data problem. Therefore, we will focus on generalizing this to higher dimensions.  

\subsubsection{Matthew's correlation coefficient in two dimensions}\label{sec:two}
The Matthews correlation coefficient is a well-known metric for the goodness of the fit for the binary classifier (for example see \cite{matthews1975comparison}).  
Formally speaking, we denote by 1 the positive class and by 0 the negative class. Matthew's correlation coefficient is merely the Pearson correlation between the target and the prediction. Doing some algebra we obtain the formula:
\begin{equation}\label{EQ:BinaryMatthew}
 MCC=\frac{TP*TN-FP*FN}{\sqrt{(RP+FP)(TP+FN)(TN+FN)(TN+FP)}} . 
\end{equation}

Next, we will rewrite (\ref{EQ:BinaryMatthew}) in a more geometric fashion. 
For understanding the geometrical meaning of the two-dimension Matthews correlation coefficient
we need the next notations. $P$ and $N$ be the set of all positive and negative receptively. Akin to this, $PT$ and $PN$ are the predicted positive and predictive negative. 
$G(\dot,\dot)$ is the geometric mean as defined in (\ref{EQ:geometric_mean}). 
 The Matthews correlation coefficient is taking the determinant of the correlation matrix of the target and the predicted \cite{jurman2012comparison}.

$P(A|B)$ is the probability of $A$ given $B$.
Consider the normalized confusion matrix
\begin{equation}\label{EQ:normalized_confusion}
\begin{pmatrix} G(P(TP|P),P(TP|PT))& G(P(TP|P),P(TN|PN))&\\
G(P(TN|N),P(TP|PT))& 
G(P(TN|N),P(TN|PN))&
\end{pmatrix} . 
\end{equation}

Then the Matthews correlation coefficient is the determinant of the matrix defined in ( \ref{EQ:normalized_confusion}). 

\begin{lemma} 
The Matthews correlation coefficient is invariant to:
\begin{enumerate}
    \item  Flipping the target and the predicted vector.
    \item Transforming the True and Negative classes. 
 \end{enumerate}
\end{lemma}
\begin{lemmaproof}
Flipping the prediction vector and the target vector the new normalized confusion matrix would be the transpose matrix of Equation (\ref{EQ:normalized_confusion}). \\
Flipping the Positive and the Negative class one would change the diagonal elements of (\ref{EQ:normalized_confusion}) and then take the transpose. \\
In both cases, the determinant value is fixed.  
% \textbf{QED}.
\end{lemmaproof}
\renewcommand\qedsymbol{$\blacksquare$} 
 
% \textbf{Proof. }
% Flipping the prediction vector and the target vector the new normalized confusion matrix would be the transpose matrix of Equation (\ref{EQ:normalized_confusion}). \\
% Flipping the Positive and the Negative class one would change the diagonal elements of (\ref{EQ:normalized_confusion}) and then take the transpose. \\
% In both cases, the determinant value is fixed. \textbf{QED}.

We note that the normalization ensures that the measurement would be invariant to scale. It is well known that the geometrical meaning of the determinant of a $2\times 2$ matrix is the area of the parallelogram spanned by vectors that form the matrix. One can conclude that if the classifier favorites one class on the top of the other then the determinant would be near zero. Moreover, if the determinant is negative this means that reverting the prediction will improve accuracy.

\section{Multi-Class Prediction}\label{Sec:multi}
In the previous section, we discuss binary classification.
The focus was on setting to fundamental basis for the multi-classes case.  
In this section, we will present methods to generalize the above to multi-classes. To be able to do that we need the following setting:
\subsection{Generalizing request }
Let's start with a review of the requirements for the generalization. 
In two dimensions the hierarchy of classes is clear. 
Nonetheless, in high dimensions, the order of classes might be blurry. Therefore, a robust classifier must be invariant for any permutation of the order of the classes.  
To formulate this we need the next definition:
\begin{definition}
Multi-class metric prediction is called \textbf{suitable metric} if it agrees with the following:
\begin{enumerate}
    \item Bounded by one and minus one. 
    \item Invariant for classes permutations.
    \item Invariant for flipping the target and the predicted. 
    \item For a perfect fit the score is one. 
    \item Only for a perfect fit the score is one.
\end{enumerate}
If the last condition fails but there exists a simple map for the perfect fit it is called
\textbf{semi- suitable metric}.
\end{definition}
 It is trivial that the accuracy metric satisfies the above. However, as we discussed one might want to choose other metrics. In the following, we will define a few. 
\subsection{Generalizing Known KPI's}

After defining the requirement of general metrics, we provide an example of such.
One approach is averaging the binary metrics between all pair possible values in the target class.
Next, we will generalize this approach and add metric oriented for multi-classes. 

\subsubsection{All against all}
Measuring the classifier with the all against all approach. Namely, averaging the score of each pair. In detail, for each pair, we are omitting the predictions and the values other than the pair. For these values, we are taking the value for some metric (say Matthews correlation coefficient).  
This approach was introduce in metrics' section in  \cite{tanha2020boosting}.
\begin{theorem}
If we are averaging with a reasonable average and the underline metric is a suitable metric then the metric is a suitable metric.
\end{theorem}
\begin{theoremproof}
Since it is a reasonable average we get that all of the properties of the underline metric are induced to the average. Since we choose the underline metric to be suitable the metric is suitable. 
\end{theoremproof}\renewcommand\qedsymbol{$\blacksquare$} 
\subsubsection{General F1-score}
For using F1-score, one needs to define a proper measurement for Precision-Sensitivity or NPV-Specificity. While in the binary case these measurements can be defined straightforwardly, It is not the case for multi-class problems. Using a one-to-many approach we can average over the harmonic means formula that was presented.
Viz, 
\[
F1_{general} = S(H_1,H_2, \ldots H_n) \quad H_i = H(P(PC_i| C_i),P( C_i|PC_i)) ,
\]Where $C_i$ are the members of the class $i$, and $PC_i$ is the predicted to the class $i$. $H$ is the Harmonic mean as defined in (\ref{EQ:harmonic_mean}). 
And $S$ is any reasonable average. 
\begin{theorem}
$F1_{general}$ is a suitable metric. 
\end{theorem} 
\begin{theoremproof}
Since probability measure is bounded by one the first condition holds. The second is since we choose a reasonable average. The third and fourth properties are proved by the definition of Sensitivity and Precision.
\end{theoremproof}
\renewcommand\qedsymbol{$\blacksquare$} 

% \textbf{Proof. } Since probability measure is bounded by one the first condition holds. The second is since we choose a reasonable average. The third and fourth are by the definition of Sensitivity and Precision. \textbf{QED}

Replacing the Harmonic mean with a geometric mean gives the general Fowlkes–Mallows index.
One can rewrite the proof of the generalized F1-score and get that the generalized Fowlkes–Mallows index is a suitable metric. In addition, using (\ref{EQ:mean_inequ}) we have $F1_{generalize}\le FM1_{generalize}$.
Following \ref{EQ:lpn} one can define the following metric 
\[\ell^p(P( i |predict i),P(predict j | j)  )\quad p \le 1, \quad 1\le i,j \le n . \]
One get a desirable metric. This is dues to the basic properties of $\ell^p$ spaces. 

% \subsubsection{Cramér's $\phi_c$}
\subsubsection{Cramér's $\phi_c$ }

One can measure the goodness of the fit as an association test. The popular choice would be the Cramér's $\phi_c$ \cite{acock1979measure}. Namely,
\[
\phi_c = \frac{\sqrt{\chi^2}}{n-1}, \quad \chi^2= \sum_{1\le i,j\le n}\frac{(n_{i,j}-n_in^*_j)^2}{\frac{n_in^*_j}{n}},
\]where $n_{i,j}$ the number of prediction $j$ of the class $i$. Moreover, $n_i$ is the size of the class $i$ of the data. $n^*_j$ is the size of the set which was predicted $j$.

It is well-known that for two dimensions Cramér's $\phi_c$ agrees with the Matthews correlation coefficient.
\begin{theorem}
Cramér's $\phi_c$ is a semi-suitable metric.
\end{theorem}
\begin{theoremproof}
 One can consider Cramér's $\phi_c$ as the goodness of the conditional probabilities of each class. Therefore, the first three properties hold. For the fourth, we achieve this by the properties of the $\chi$ function. We note that for two dimensions this measure agrees with the Matthews correlation coefficient.   
\end{theoremproof}
\renewcommand\qedsymbol{$\blacksquare$} 

\subsubsection{ General Matthews correlation coefficient}

Detecting the bias of a predictor had gained momentum in recent years. 
E.g., the analysis of fair AI. (for deep discussion on the subject see \cite{o2016weapons}).
 Nonetheless, one can point out that the Matthews correlation coefficient does not suffer from this problem. This leads us to generalize this metric to multi-dimensions for $n$ classes.

For this, we are defining the generalized normalized confusion matrix $M$. Let the $M$ be the $n\times n$ matrix,
\begin{equation}\label{EQ:generalized_conf}
M_{i,j} =G\left(P( i |predict j),P(predict j | i)\right) \quad 1\le i,j\le n  . 
\end{equation}
Recollect that $G(\dot,\dot)$ is the geometric mean as defined in (\ref{EQ:geometric_mean}). Since the entries in the matrix above are probabilities they are scale-invariant. Hence, the matrix is scaled invariant. Moreover, one can point out that matrix $M$ is symmetric. Thus, all of the eigenvalues are reals. However, some of them might be negative.  The geometric meaning of taking the determinant is the hyper volumes of the hyper-parallelogram.

\begin{figure}[p]
    \centering

    \includegraphics[width=12cm]{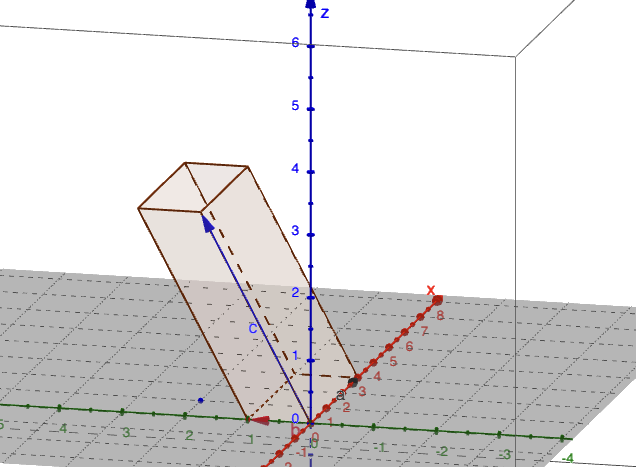}
    \caption{The Parallelepiped which is spanned by the vectors a=(3,0,0),b=(0,1,0), c=(1,2,3)}
    \label{fig:galaxy}
\end{figure}

We need to know to show that the determinant of the matrix $M$ reflects well the prediction's performance. We will show this using the following theorem

\begin{lemma}\label{lemma:one} 
The determinant's magnitude of the $M$ is bounded by $1$. In addition, the value $1$ is attained only for the perfect test.
\end{lemma}
\\
% \textbf{Proof. }
For proving this theorem we will need three steps.   

 The first claim  states the following:
\begin{claim}\label{step1}
 Let $\{a_i\}_{i=1}^n$ be a set of vectors such that each entry is non-negative ($a_i^j \ge 0$). Then the volume of the paralleled which is generated by this vector is smaller or equal to the product of the   $\{a_i\}_{i=1}^n$ , where equality is obtained only if the vectors generate a permutation matrix
\end{claim}
  
 The observation for this claim is obtained from linear algebra: Let $\{a_i\}_{i=1}^n$ be a set of vectors such that each entry is non-negative ($a_i^j \ge 0$) then
\[volume\left(parallelepiped(\{a_i\}_{i=1}^n)\right) \le \prod_i^n \left(\sum_j^n a_i^j\right) = \prod_i^n \ell_1( a_i),  \]

where $volume\left(parallelogram(\{a_i\}_{i=1}^n)\right)$ is the volume of the parallelepiped spanned by the vectors $\{a_i\}_{i=1}^n$. $\ell_1(v) =\sum_j^n|v_j|$. 

\begin{claimproof}
We prove it by induction. Without loss of generality, we assume that the rows are arranged in decreasing order of the sum.\\ The initial step is trivial. Assume this is true for $k$ steps, the $k+1$ vector can be written as the sum of two vectors. The first is in the parallelogram and the second is perpendicular to it. The hypervolume would be the hypervolume reached on the $k$ step times the perpendicular vector sum. Since this is bounded by the sum of the interior vector the hypervolume is bounded by it. As we claim equality holds only when the vectors are orthogonal. In the case of the vector that forms the normalized confusion matrix, the non-negativity of the values implies that the vector contains only the values zero and one. This proves this step. 
\end{claimproof}
% \qedsymbol{} 
 
\renewcommand\qedsymbol{$\blacksquare$} 
% We prove it by induction. Without loss of generality, we assume that the rows are arranged in decreasing order of the sum.\\ The initial step is trivial. Assume this is true for $k$ steps, the $k+1$ vector can be written as the sum of two vectors. The first is in the parallelogram and the second is perpendicular to it. The hypervolume would be the hypervolume reached on the $k$ step times the perpendicular vector sum. Since this is bounded by the sum of the interior vector the hypervolume is bounded by it. Equality holds only when the vectors are orthogonal. In the case of the vector that forms the normalized confusion matrix, the non-negativity of the values implies that the vector contains only the values zero and one. This proves this step. 
 
 The second step is the following claim:
 \begin{claim}\label{step2}
 The sum of all of the entries of matrix $M$ is bounded by $n$
\end{claim}
 
 The motivation is obtained by using the inequality of means. This yields that for every entry, in the matrix, is bounded by an average of entries of left and right stochastic matrices. Since for each stochastic matrix, the sum of the entries is $n$ this average is $n$ as well.
 
 \begin{claimproof}
 Let therefore the mean in-equality (\ref{EQ:mean_inequ}):
\[
G\left(P( i |predict j),P(predict j | i) \right)\le \frac{ P( i |predict j) +
P(predict j | i)}{2} . 
\] 
This yields, 
 \begin{eqnarray}\sum_{j}^n \sum_{i}^n G\left(P( i |predict j),P(predict j | i)\right)  &\le& \\ \frac{ \sum_{j}^n\sum_{i}^n
P( i |predict j) +
\sum_{i}^n\sum_{j}^n P(predict j | i)}{2} &=&\\
\frac{ \sum_{j}^n 1 +
\sum_{i}^n 1}{2}&=& n .
 \end{eqnarray} 

\end{claimproof}
\renewcommand\qedsymbol{$\blacksquare$} 

%  Let therefore the mean in-equality (\ref{EQ:mean_inequ}):

% \[
% G(P( i |predict j),P(predict j | i) \le \frac{ P( i |predict j) +
% P(predict j | i)}{2} . 
% \] 

% This yields, 
%  \begin{eqnarray}\sum_{j}^n \sum_{i}^n G(P( i |predict j),P(predict j | i)  &\le& \\ \frac{ \sum_{j}^n\sum_{i}^n
% P( i |predict j) +
% \sum_{i}^n\sum_{j}^n P(predict j | i)}{2} &=&\\
% \frac{ \sum_{j}^n 1 +
% \sum_{i}^n 1}{2}&=& n .
%  \end{eqnarray} 
The last step is the following:
\begin{claim}\label{step3}
If $\sum_i^n x_i \le n$ and $ x_i\ge 0$ then the maximum of the product $\prod_i^n x_i$ is 1. Furthermore, it is obtained only when all the  $ x_i$ are equal. Moreover, the determinant is bounded by this product 

\end{claim} 
\begin{claimproof}
Using the mean in-equality (\ref{EQ:mean_inequ}) we have 
\[\sqrt[n]{\prod_i^n x_i} \le \frac{\sum_i^n x_i}{n} \le 1. \]
For this we have  \[
\sqrt[n]{\prod_i^n x_i} \le 1 \quad \Rightarrow{} \quad 
\prod_i^n x_i \le 1 . \] Equality holds only when all of the numbers are equal to one.

Taking the Steps \label{step1}, \label{step2} and \label{step3} we have \begin{eqnarray}
det(M) &=& volume\left(parallelepiped(\{a_i\}_{i=1}^n)\right) \\ &\le& \prod_j^n \left( \sum_i^n G(P( i |predict j),P(predict j | i)\right)\\ &\le& 1 . 
 \end{eqnarray}

\end{claimproof}
\renewcommand\qedsymbol{$\blacksquare$}\\  
We can now combine the three claims for proving the lemma above:
\begin{lemmaproof}
We learn from step two that the sum of all of the entries of the matrix $M$ is bounded by $n$. By step \label{step3} we have that the determinant is bounded above by the product of the sum of the rows of $M$. Using step \label{step1}, this product is 1 only for permutation matrix. Which reflect perfect results.
\end{lemmaproof}
\renewcommand\qedsymbol{$\blacksquare$}  
% The last step is showing that if $
% \sum_i^n x_i \le n$ and $ x_i\ge 0$ then using the mean in-equality (\ref{EQ:mean_inequ}) we have 
% \[\sqrt[n]{\prod_i^n x_i} \le \frac{\sum_i^n x_i}{n} \le 1. \]
% For this we have  \[
% \sqrt[n]{\prod_i^n x_i} \le 1 \quad \Rightarrow{} \quad 
% \prod_i^n x_i \le 1 . \] Equality holds only when all of the numbers are equal to one.

% Taking the three steps we have \begin{eqnarray}
% det(M) &=& volume(parallelogram(\{a_i\}_{i=1}^n)) \\ &\le& \prod_j^n ( \sum_i^n G(P( i |predict j),P(predict j | i)))\\ &\le& 1 . 
%  \end{eqnarray}

% \textbf{QED}

\begin{lemma}\label{lemma:inv_mat}
The generalized Matthews correlation coefficient is invariant for class renaming the classes and flipping the target and the predicted. 
\end{lemma}
\begin{lemmaproof}
As above flipping, the target and the perdition would mean taking the transpose. 
Renaming the classes means taking a permutation on the columns and then taking the same permutation on the rows. For an even permutation, the determinant is fixed. For an odd permutation, the determinant changes the sign of the determinant over the rows permutation. Yet, the column permutation changes the sign back.
\end{lemmaproof}
\renewcommand\qedsymbol{$\blacksquare$} 
\\

The next theorem is straightforward from the above, 
\begin{theorem}
The General  Matthews correlation coefficient is a semi-suitable metric.
\end{theorem} 
The proof is the combining Lemmas \ref{lemma:one}, \ref{lemma:one}.

If the predictor disfavors one class, the hyper-volume would be of one dimensionless. Thus, it would be small. Namely, the predictor not only measures accuracy but it detects the bias of the classifier.  
If the determinant is $1$ then there exists an even permutation of the classes. If the value is $-1$ there exists an odd permutation. 
Example of non-perfect prediction with value $1$:

Consider a classifier with the following normalized generalized confusion matrix,
\[
\begin{pmatrix} 0& 0&1 \\
1&0&0 \\ 
0&1&0
\end{pmatrix} . 
\]The determinant is one but the prediction is zero. Nevertheless, applying the transformation 
\[class_1 \longrightarrow class_2 \longrightarrow class_3\longrightarrow class_1 \]
would give the perfect prediction is straight forward.

We note that this prediction method is based on probabilities. Therefore, using prior methods, e.g., Additive (Laplace) smoothing \cite{chen1996empirical},  is straightforward. 
Another advantage of this approach is that if one class gets less prediction the volumes of the parallelogram would plummet. Thus, the user would not use this predictor. 
For example, consider the generalized confusion matrix
\[
\begin{pmatrix} 20& 6&0 \\
2&20&0 \\ 
12&12&8
\end{pmatrix} . 
\]
One can point out that the classifier prefers to avoid the third class. Following this, we should grade poorly the classifier. Indeed, the score here is 0.235. This is since the parallelogram is almost flat.

Taking any reasonable average of the diagonal elements of the generalized normal matrix to define in \ref{EQ:generalized_conf} one would get the generalized Fowlkes–Mallows index. A natural example of this is taking the trace and dividing it by the number of classes.
Additional insight is that if one replaces the Geometric mean with any reasonable average which is dominant by the Algebraic mean, i.e., $S(a,b) < 
A(a,b)$ for all $a\ne b$, then the above would hold.

\section{The code}\label{sec:code}
The following is the python code. We are using Python script with Numpy \cite{numpy} ,Scikitlearn \cite{scikit-learn}  and 
Pandas\cite{reback2020pandas}. The source can be found \href{https://github.com/natank1/UriProjects}{here}

\begin{lstlisting}
import pandas as pd
import numpy as np
import scipy as sc


hm = sc.stats.mstats.hmean
gm = sc.stats.mstats.gmean

def ratio_mat(mat,average):
    m1 = mat/mat.sum(axis = 1)
    m2  = mat/mat.sum() 
    
    return  [[average([m1.loc[i][j],m2.loc[j][i]])
            for i in m1.index] for j in m1.columns]

def genralize_f1(mat,average_2 = hm):
    H_mat = ratio_mat(mat,hm)
    return average_2([H_mat[i][i] for i in range(len(mat))])

def genralize_mat(mat):   
    G_mat = ratio_mat(mat,gm)
    return np.linalg.det(G_mat)

def Cramer_phi(mat):
    X2 = sc.stats.chi2_contingency(mat, correction=False)[0]
    n = np.sum(np.sum(mat))
    minDim = min(mat.shape)-1
    return np.sqrt((X2/n) / minDim)

a = [[5,6,2],[2,8,11],[8,2,10]]
mat = pd.DataFrame(data = a)
print(genralize_mat(mat))
print(genralize_mat(pd.DataFrame(data =
                           np.identity(4))))
print(genralize_f1(mat))
print(genralize_f1(pd.DataFrame(data =
                           np.identity(4))))
print(Cramer_phi(mat))
print(Cramer_phi((pd.DataFrame(data =
                           np.identity(4)))))
\end{lstlisting}

\section{Closing remarks}\label{sec:clos}
In this paper, we suggest a method for generalizing the goodness of fit metrics for multi-classes. In particular, we suggest the Generalize F1-score and the generalized Matthews correlation coefficient into $n$ classes. This adds to the known Cramér's $\phi_c$ metric. Both generalizations are geometrically oriented. Thus, renaming the classes or the predicted vector is invariant. Moreover, we supply a geometric meaning of the metrics. 
In this paper, we do not assume any topological structure of the target classes. Moreover, no prevalence is assumed. 
\begin{itemize}
 \item We note that both Cramér's $\phi_c$ and the generalized Matthews correlation coefficient extend the binary Matthews correlation coefficient. Each one of them takes a different interpretation of this metric. 
 \item Metrics generalization for the goodness of fit in geometric tools seems a promising area of research. The authors strongly encourage the readers to continue to study this field.
 \item We realized that many of our steps used stochastic matrices manners. A future objective will be studying whether this topic can be fully studied using this domain.

\item The binary case was derived from the Pearson matrix. Using the naive approach,  multi-class problems distort this approach due to the lack of Pearson score for more than two dimensions. One can consider the classes as integers. However, this would lead him to search for linear relations in a non-Euclidean space. This might differ from the nature of the setting. Thus, we fill that this direction is indeed a future challenge.
 
\end{itemize}

% We note that both Cramér's $\phi_c$ and the generalized Matthews correlation coefficient extend the binary Matthews correlation coefficient. Each one of them takes a different interpretation of this metric. 

% Metrics generalization for the goodness of fit in geometric tools seems a promising area of research. The authors strongly encourage the readers to continue to study this field.

% We realized that many of our steps used stochastic matrices manners. A future objective will be studying whether this topic can be fully studied using this domain 

% The binary case was derived from the Pearson matrix. Using the naive approach,  multi-class problems distort this approach as we don't have a Pearson score for more than two dimensions. We can consider the classes as integers but then the outcome is order dependent and if we manipulate Pearson we will find ourselves searching for linear relations in a non-Euclidean space. We fill that this direction is indeed a future challenge 

\section*{Acknowledgement}
I offer my sincerest gratitude to our dear friends: Shlomo Yona, Nir Sharon, and Tal Galili. Without their help, I would never do this research. 

\printbibliography
\end{document}